\documentclass[conference]{IEEEtran}
\IEEEoverridecommandlockouts
\usepackage{amsmath,amssymb,amsfonts}
\usepackage{algorithmic}
\usepackage{graphicx}
\usepackage{xcolor}
\usepackage{subfig}
\usepackage{multirow}
\usepackage{booktabs}
\usepackage{hyperref}
\usepackage{float}

\usepackage[nomain, toc, acronym]{glossaries}

\newacronym{dl}{DL}{Deep Learning}
\newacronym{ad}{AD}{Anomaly Detection}
\newacronym{ml}{ML}{Machine Learning}
\newacronym{ttl}{TTL}{Time-to-Live}
\newacronym{ids}{IDS}{Intrusion Detection System}
\newacronym{mlp}{MLP}{Multilayer Perceptron}
\newacronym{relu}{ReLU}{Rectified Linear Unit}
\newacronym{ip}{IP}{Internet Protocol}
\newacronym{ffnn}{FNN}{Feedforward Neural Network}
\newacronym{fcnn}{FCNN}{Fully Connected Neural Network}

\begin{document}

\title{EagerNet: Early Predictions of Neural Networks for Computationally Efficient Intrusion Detection}

\author{\IEEEauthorblockN{Fares Meghdouri, Maximilian Bachl and Tanja Zseby}
\IEEEauthorblockA{TU Wien\\
Vienna, Austria\\
firstname.lastname@tuwien.ac.at}}

\IEEEoverridecommandlockouts
\IEEEpubid{\makebox[\columnwidth]{978-0-7381-4292-0/20/\$31.00~\copyright2020 IEEE \hfill} \hspace{\columnsep}\makebox[\columnwidth]{ }}

\maketitle

\IEEEpubidadjcol

\begin{abstract}

\glspl{fcnn} have been the core of most state-of-the-art \gls{ml} applications in recent years and also have been widely used for \glspl{ids}. Experimental results from the last years show that generally deeper neural networks with more layers perform better than shallow models. Nonetheless, with the growing number of layers, obtaining fast predictions with less resources has become a difficult task despite the use of special hardware such as GPUs. We propose a new architecture to detect network attacks with minimal resources. The architecture is able to deal with either binary or multiclass classification problems and trades prediction speed for the accuracy of the network. We evaluate our proposal with two different network intrusion detection datasets. Results suggest that it is possible to obtain comparable accuracies to simple \glspl{fcnn} without evaluating all layers for the majority of samples, thus obtaining early predictions and saving energy and computational efforts.
\end{abstract}


\section{Introduction}
\glspl{fcnn} have gained remarkable attention in various domains in recent years due to the availability of data and computational resources for research. Many state-of-the-art architectures currently used in different applications are effectively based on simple \gls{fcnn} architectures. Since the size of a neural network cannot be mathematically defined, experts use rule-of-thumb to define the best architecture that fits their needs. Yet, experiments show that the more layers (and neurons) the network includes, the better its learning performance becomes \cite{eldan_power_2016}.

\begin{figure}
  \includegraphics[width=\linewidth]{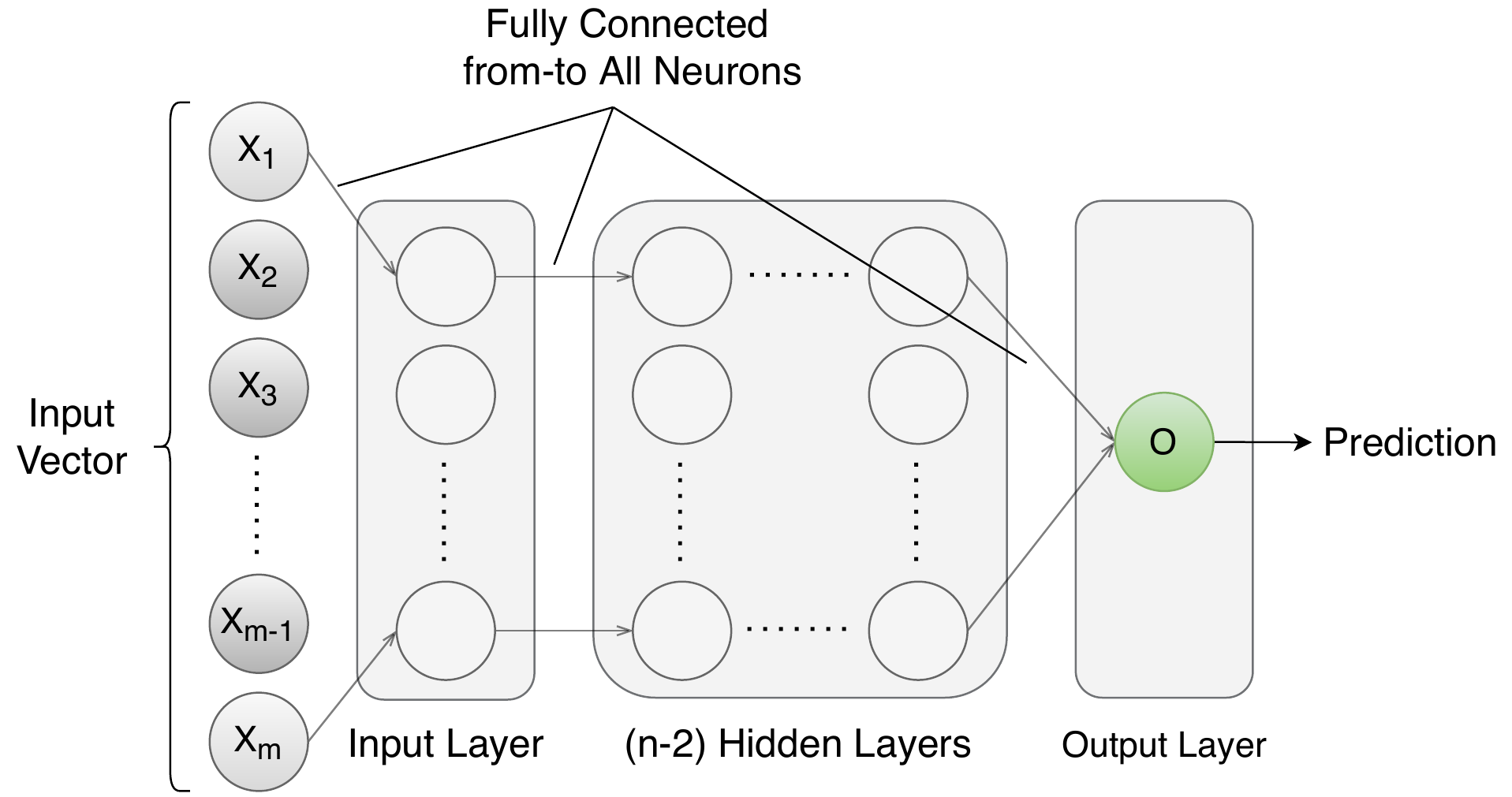}
  \caption{Conventional \gls{fcnn} architecture.}
  \label{fig:fcnn}
\end{figure}

Traditional \gls{fcnn} are based on simple logistic regression units that combine multiple inputs multiplied by a set of weights and passed through an activation function (e.g. sigmoid \cite{noauthor_sigmoid_2020}) to obtain a scalar which for the sake of consistency we call a prediction. The weights are adjusted to give the closest possible output to a ground-truth. As far as neural networks are concerned, logistic regression units are stacked together to create layers and then layers are stacked together again to create the network itself. \autoref{fig:fcnn} demonstrates a basic \gls{fcnn} that takes an input vector $\textbf{x}^{T} = \lbrace x_{1}, x_{2} ... x_{m} \rbrace$ and generates a prediction. Adjusting weights in such situations is a non-trivial problem since the weights of each layer may have an impact on the following layers. Fortunately, several algorithms have been proposed to reduce the complexity of the task (e.g. gradient descent learning) while achieving the same goal: reduce the "gap" (loss) between predictions and actual desired output. In practice, larger networks consist of more neurons and thus more weights, allowing more data patterns to be learned.

In terms of efficiency, two problems arise when using an extremely deep neural networks: (1) the loss during training needs more time to converge and (2) the time necessary to make a prediction is proportional to the number of layers. Even though most modern hardware is able to compute predictions without significant delay, prediction speed is still relevant for learning problems, in which many predictions are required per unit of time and delay is important. This is, for example, relevant in the context of object detection in computer vision or also network intrusion detection. 

To obtain early predictions, we propose a new architecture that, on the one hand, uses the entire network capacity for learning network intrusions and, on the other hand, stops the forward-pass and makes predictions as soon as confidence reaches a certain threshold. In other words, it makes it possible to avoid evaluating the entire network with all layers for some samples and to stop the neural network evaluation at a particular layer if the achieved confidence at that point is high enough. We call different \glspl{fcnn} architectures that uses the same proposed approach \emph{EagerNets} (Eager Stopping Networks). EagerNets therefore allow, where possible, the reduction of computing resources and energy usage while achieving detection performance comparable to that of a full forward-pass.

In the following sections, we present the existing related work, introduce the concept behind EagerNet and finally show and analyze the evaluation results based on two network intrusion detection datasets.

\section{Related Work}

The concept of \textit{cascading} classifiers encompasses several classifiers that are used in a cascade (like a chain) \cite{viola_robust_2004, xu_classifier_2014, wang_efficient_2015}. If accuracy of a classifier in the beginning is good enough, then there is no need to evaluate the other classifiers in the chain (cascade), which saves computational resources. This approach was predominantly explored for classical machine learning methods. Two possibilities are to either place all classifiers in a row or in the shape of a tree. If a tree is used, different classifiers can be used at each step, depending on which features are probably relevant for the sample currently under investigation. The difference to our approach is that we exclusively consider neural networks as the classifier and that our approach is significantly more straightforward to implement.

\cite{bolukbasi_adaptive_2017} aim to reduce the computational complexity during evaluation for large Convolutional Neural Networks (CNNs) using an additional classifier after each layer. These additional classifiers learn to look at the output of the previous layer and decide whether it's necessary to continue evaluating the next layers. Unlike our approach, this approach is primarily geared towards making the evaluation of a given neural network more efficient, while our approach aims to train a neural network from scratch. Furthermore, their approach requires advanced math, making it difficult to implement and deploy for practitioners.

The approach that is most similar to ours is presented by \cite{leroux_resource-constrained_2015,leroux_cascading_2017}. They introduce an additional output layer after each layer and then decide whether the next layers should be evaluated based on whether the confidence of the current output layer is high enough. They also argue that if many layers are required, the computation can be offloaded to a more powerful machine in the cloud, which saves further computational effort. The main difference of this approach compared to ours is that we can show that no additional output layers are needed but instead our approach only needs one extra neuron per layer, which further reduces the computational overhead of our approach.

Another line of work \cite{seo_neural_2018,yu_learning_2017, graves_adaptive_2017, bachl_sparseids_2020} focuses on teaching Recurrent Neural Networks (RNNs) -- neural networks for sequences -- to skip irrelevant parts of the input sequence.

The major difference between all existing works and ours is simplicity in terms of implementation complexity and interpretability. While many of the previously proposed works report good results, they usually require advanced math knowledge to be implemented, which hinders their deployment in real-world scenarios. These also often contain components such as additional output layers, which add computational overheads that are shown not to be necessary.

To the best of our knowledge, our work considers for the first time such an approach for network stream data to obtain early flow classification yet with remarkable performance.

\section{Supervised Learning for Network Intrusion Detection}


\gls{ml} has been widely used in the last decade for network traffic analysis and specifically anomaly detection systems. Many of the works proposed \gls{ids} architectures based on well-known supervised techniques \cite{buczak_survey_2016,berman_survey_2019}. The concept of having a classifier trained on pre-stored attack patterns and using it to predict similar behaviours is commonly used whereby remarkable success has so far been achieved \cite{vigneswaran_evaluating_2018,ferrag_deep_2020,almseidin_evaluation_2017}. In particular, \glspl{fcnn} are commonly utilized whenever large amounts of network traffic are available. In order to present network data to such architectures, a number of traffic representations have been proposed depending on the application. The statistical representation of flow characteristics is so far the most widely used. In this work, we define a network flow to be the exchange of packets between two end-hosts. Packets can be identified and aggregated using the five-tuple key: \emph{sourceIP}, \emph{destinationIP}, \emph{sourceTransportPort}, \emph{destinationTransportPort} and \emph{protocolIdentifier}. Thereafter, packet header features are extracted and statistical combinations are computed. \autoref{tab:features} shows the CAIA \cite{williams_preliminary_2006} network traffic representation. It consists of 12 features, 7 of which are measured in both directions and 2 of which are expanded using four statistical combinations (i.e. \emph{A} becomes \emph{mean(A)}, \emph{min(A)}, \emph{max(A)} and \emph{stdev(A)}). We use CAIA for all experiments in this paper and set an observation timeout of 1,800 seconds (slightly longer than the longest attack activity in the dataset), after which we terminate flows. Each flow is considered to be one data sample and has two sets of labels: a binary (attack/benign) label and also the attack family (one binary label for each attack family using one-hot-encoding). The ultimate goal is that after a training process, the trained \gls{ids} can correctly classify new flows and also determine what kind of attack family they belong to.

\begin{table}[ht!]

	\centering
	\caption{CAIA flow representation.}
	\label{tab:features}

	\footnotesize
	\begin{tabular}{|c|c|c|}
		\hline
		\textbf{Direction}                    & \textbf{Features}        & \textbf{Statistical Operations}        \\
		\hline

		                                      \multirow{5}{*}{No direction} & flowDurationMilliseconds & \multirow{5}{*}{None} \\
		                                      & sourceTransportPort      &                                        \\
		                                      & destinationTransportPort &                                        \\
		                                      & protocolIdentifier       &                                        \\
		                                      & octetTotalCount          &                                        \\
		\hline
		\multirow{7}{*}{\shortstack{Forward and\\ Backward}} & ipTotalLength            & \multirow{2}{*}{Mean, Min, Max, Stdev} \\
		                                      & interPacketTimeSeconds   &                                        \\
		\cline{2-3}
		                                      & packetTotalCount         & \multirow{5}{*}{None}                                       \\
		                                      & tcpSynTotalCount         &                                        \\
		                                      & tcpAckTotalCount         &                                        \\
		                                      & tcpFinTotalCount         &                                        \\
		                                      & tcpCwrTotalCount         &                                        \\

		\hline

	\end{tabular}

\end{table}

\section{EagerNet}

\begin{figure*}[htp]
\center

\subfloat[EagerNet: Forward-pass.]{%
  \includegraphics[clip,width=0.5\textwidth]{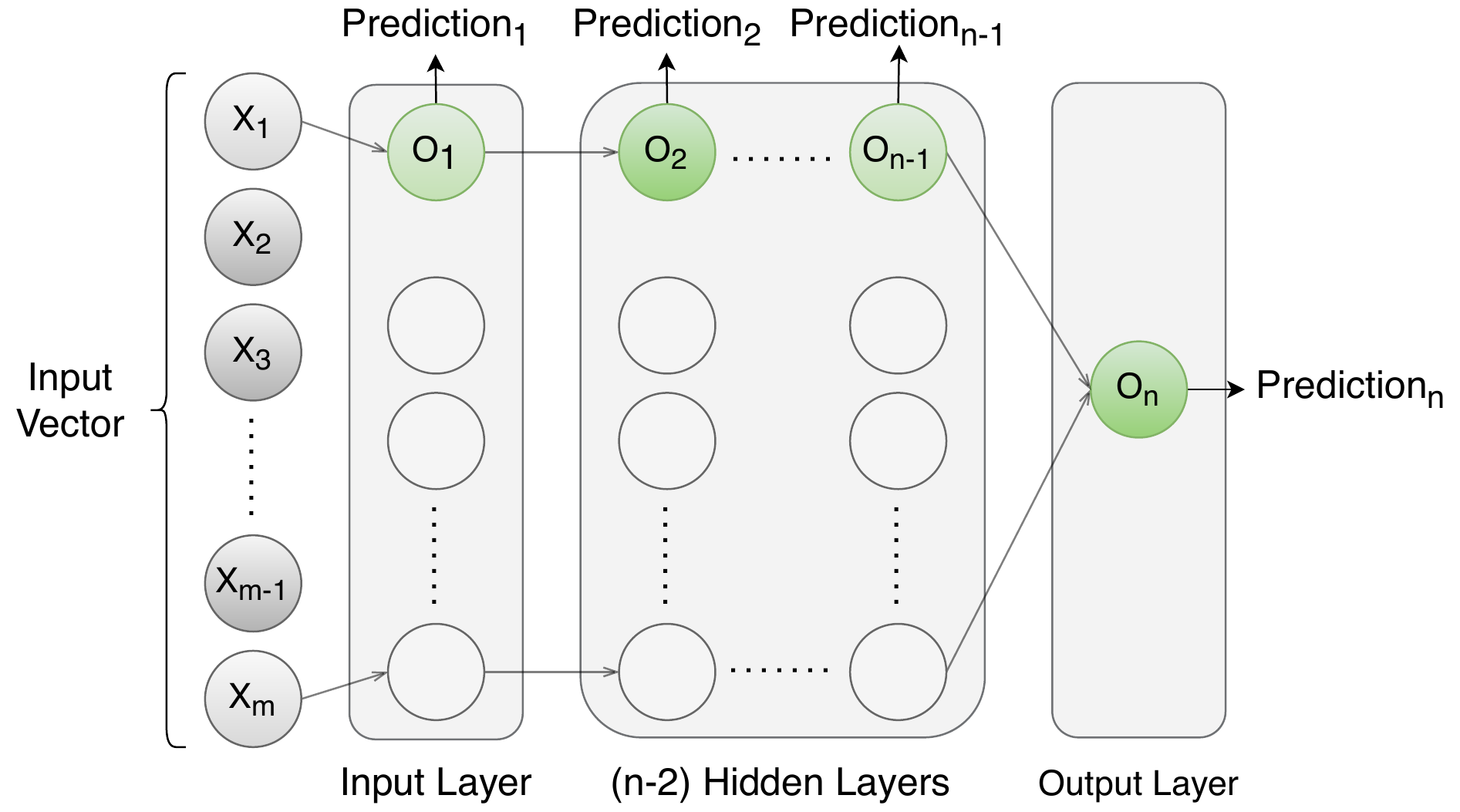}%
  \label{fig:eagerNet_forward}
}
\quad
\subfloat[EagerNet: Back-propagation of gradients.]{%
  \includegraphics[clip,width=0.42\textwidth]{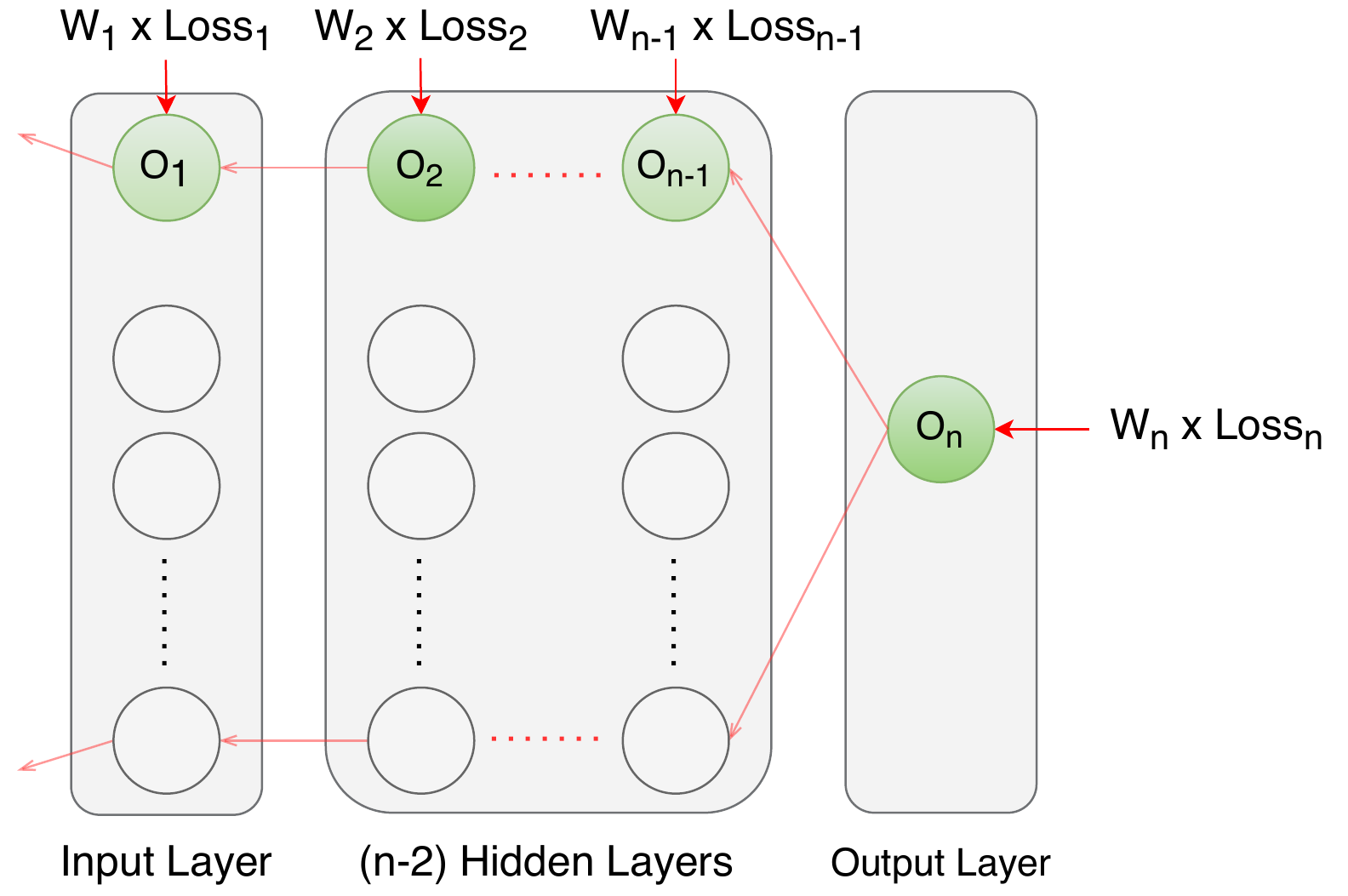}%
  \label{fig:eagerNet_backward}
}

\caption{The difference between conventional neural networks and our proposed architecture.}

\end{figure*}


Because some network attack families are simpler to detect compared to others we modify the standard \gls{fcnn} architecture so that for easily classifiable samples the network doesn't have to evaluate all layers. Our novel architecture is based on the assumption that deep neural networks with more layers can learn increasingly complex functions which in turn are only required for the classification of some particular notoriously hard-to-classify samples. As a result, we build a network such that an additional set of neurons are connected to each layer (a copy of the output neurons instead of an entire meta-network), allowing for direct predictions at each layer. The EagerNet architecture is shown in \autoref{fig:eagerNet_forward}. Initial neurons are shown in gray and output neurons (per layer) are shown in green. Once a network flow is observed and features are extracted, a data sample $\textbf{x}$ is fed to the network via the input layer. The neural network proceeds evaluating the layers one by one and yields a prediction confidence value at each layer. We define the confidence as the value that represents how sure the network is about a certain decision or sample belonging to a certain class. It is a number between 0.5 and 1 that is obtained by applying the sigmoid function \cite{noauthor_sigmoid_2020} on the output neuron and computing how far the result is from the farthest label (e.g. if the prediction scalar of a sample is 0.2, the farthest label is 1 and therefore a confidence of 0.8 that the sample belongs to class 0). By looking at the confidence the neural network determines whether additional layers should be processed or if the currently obtained confidence is high enough. This methodology ensures that simpler samples are classified early in the forward-pass and that more complicated samples are passed into deeper layers. In the section \ref{subsubsec:confidence-speed-tradeoff}, we show that EagerNet reduces resource consumption on average, while achieving comparable performance compared to normal \gls{fcnn}.

\subsubsection{Network Architecture}
For our experiments, we use a neural network that consists of as many layers as needed with Leaky ReLU activations \cite{maas_rectifier_2013} ($\alpha = 0.1$) after each layer. Moreover and to reduce over-fitting, we use neurons dropout ($r=0.2$).

\subsubsection{Binary vs. Multiclass Classification}
We experiment with two different architectures. The first architecture, consists of a binary attack detector, i.e. a 1 is an attack flow and a 0 is a benign flow. The predictions in this case are given by a single output neuron followed by a sigmoid activation function to force values between 0 and 1. During back-propagation, we use a binary cross-entropy loss and average the loss over the batch. \autoref{eq:bceloss}. shows how the loss is determined for a batch of size $N$, with the $n^{th}$ input being $x_{n}$ and the sigmoid activation function $\sigma(\cdot)$.

\begin{equation}
\label{eq:bceloss}
l = - \frac{1}{N} \sum_{n}^{N} [y_n \cdot \log \sigma (x_{n}) + (1 - y_n) \cdot \log (1 - \sigma (x_{n}))]
\end{equation}

For the multiclass approach, on the other side, there is one output neuron per attack family, and each neuron's output indicates how likely it is, that the currently processed sample belongs to the given attack family. During back-propagation, we use the categorical cross-entropy loss to penalize all outputs together, so there can be no overlap between classes. \autoref{eq:cceloss}. shows how the loss is determined for a batch of size $N$, with the $n^{th}$ input $x_{n}$ and the softmax activation function \cite{noauthor_softmax_2020} $\phi(.)$, $C$ being the number of classes or output neurons and $i$ is the current class/output.

\begin{equation}
\label{eq:cceloss}
l = - \frac{1}{N} \sum_{n}^{N} \sum_{i}^{C} y_{n,i} \cdot \log (\phi(x_{n})_{i})
\end{equation}

After computing the losses, we use $Adam$ optimizer \cite{kingma_adam_2017} for computing gradients with a learning rate of 0.001.

\subsubsection{Combined Back-propagation Loss}
Conventional \gls{ffnn} uses the loss of the last layer to adjust the gradients. In this research, and since we have multiple outputs; one at each layer, we aggregate the losses and back propagate them. \autoref{fig:eagerNet_backward} demonstrates how losses propagate back from the respective layer only. As a consequence, the weights of the last layer are affected only by the loss of the last output. Similarly, the weights of the first layer are affected by the losses of the entire network. In addition, we introduce a weighting policy for losses. We allocate a weight to each loss that is predefined prior to training. The aim is to help the network determine on which layer to optimize more. Three types of weight sets are defined:
\begin{itemize}
\item Uniform weights. all losses have the same weight:
\begin{equation}
\mathcal{W_U} = \lbrace W_{0}, ..., W_{n} \rbrace = \lbrace \frac{1}{\sum_{i=0}^{n} 1}, ..., \frac{1}{\sum_{i=0}^{n} 1} \rbrace
\end{equation}

\item Increasing weights. losses are increasingly important from first to last layer.
\begin{equation}
\mathcal{W_I} = \lbrace W_{0}, ..., W_{n} \rbrace = \lbrace \frac{1}{\sum_{i=0}^{n} i+1}, ..., \frac{n+1}{\sum_{i=0}^{n} i+1} \rbrace
\end{equation}

\item Decreasing weights. losses are decreasingly important from first to last layer.
\begin{equation}
\mathcal{W_D} = \lbrace W_{0}, ..., W_{n} \rbrace = \lbrace \frac{n+1}{\sum_{i=0}^{n} i+1}, ..., \frac{1}{\sum_{i=0}^{n} i+1} \rbrace
\end{equation}

\end{itemize}

We show the effect of the weight distributions in \autoref{tab:weights} and discuss results in \autoref{evaluation_and_discussion}.

\section{Evaluation and Discussion}
\label{evaluation_and_discussion}
In this section we discuss the different experimental results and observations. Overall, EagerNet shows promising performance that are comparable to \glspl{fcnn} that always use all layers yet with the ability to save resources. The interested reader can reproduce experiments with the provided sources in our repository\footnote{\url{github.com/CN-TU/ids-backdoor/tree/eager}}.

\subsection{Datasets}

\begin{figure}[H]
  \centering
  \begin{tabular}{@{}c@{}}
    \includegraphics[width=.8\linewidth]{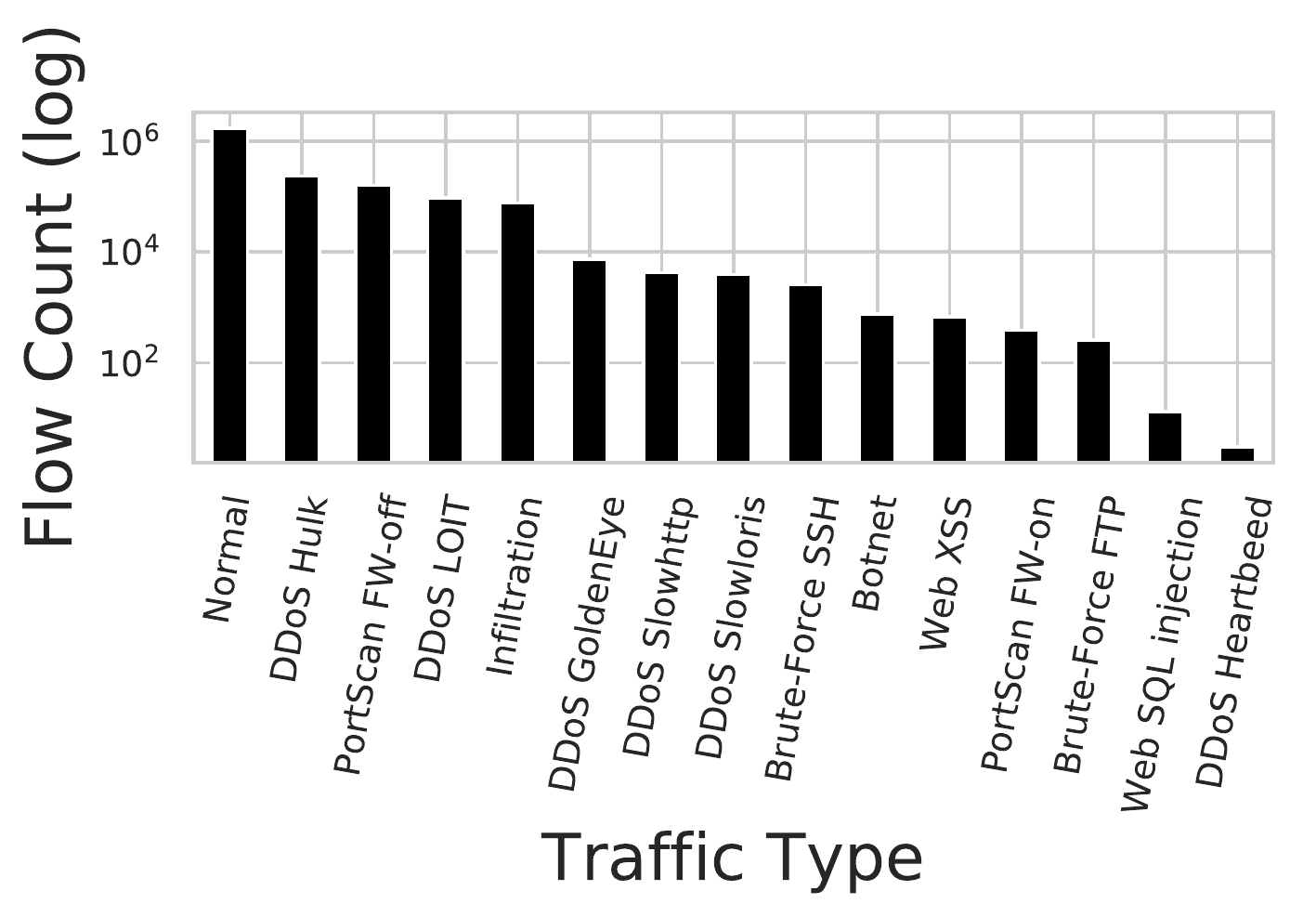} \\[\abovecaptionskip]
    \small (a) CICIDS2017
  \end{tabular}

  \vspace{0.2cm}

  \begin{tabular}{@{}c@{}}
    \includegraphics[width=.8\linewidth]{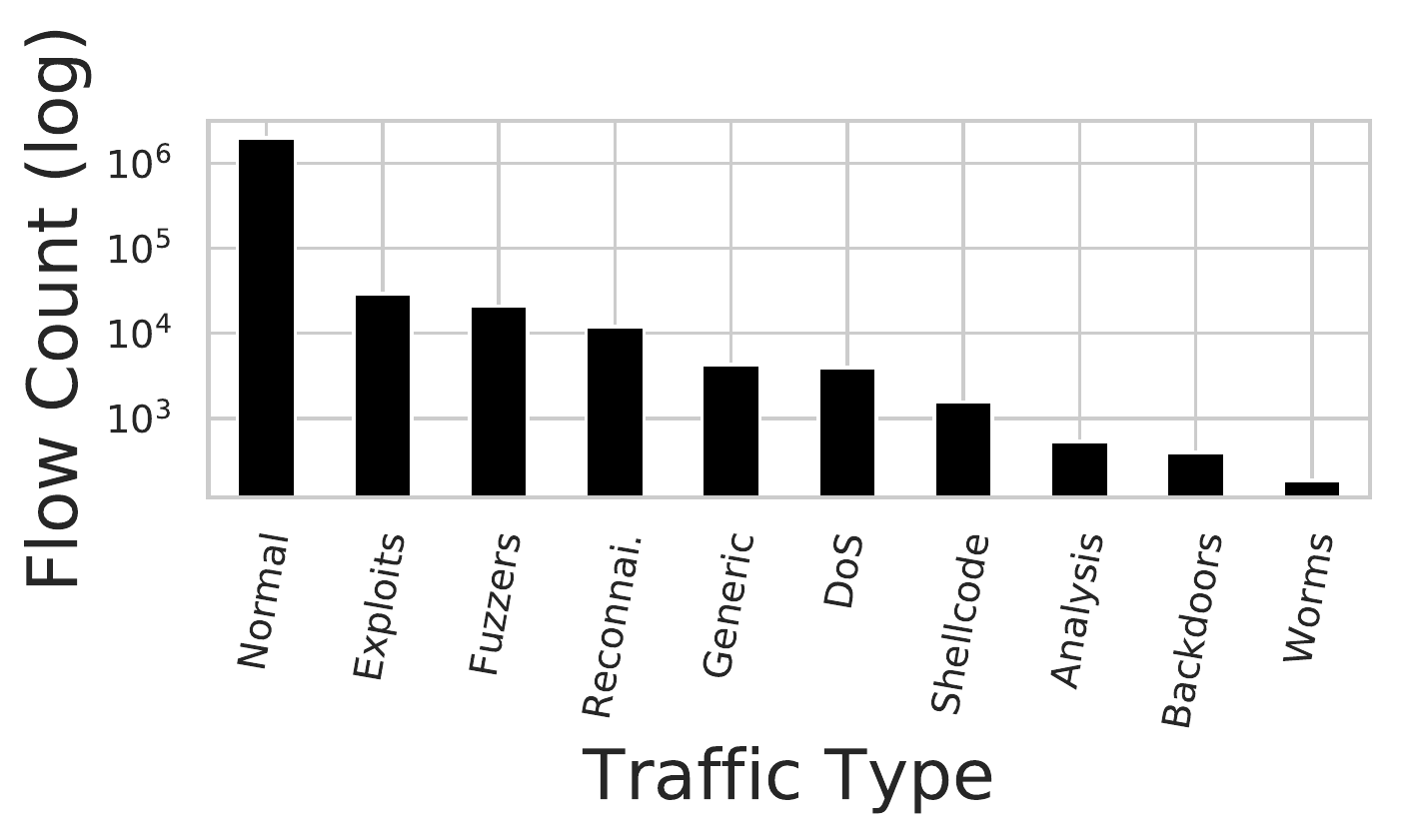} \\[\abovecaptionskip]
    \small (b) UNSW-NB15
  \end{tabular}

  \caption{Traffic distribution represented by number of flows.}
  \label{fig:datasets}
\end{figure}

For our experiments, we use two intrusion detection datasets. CICIDS2017 \cite{sharafaldin_toward_2018} which consists of 14 network attack classes in addition to realistic user profiles. UNSW-NB15 \cite{moustafa_unsw-nb15_2015} which, in addition to synthetically generated normal traffic again contains 9 attack families. As aforementioned, we represent network data utilizing the CAIA feature vector and extract 2,317,922 and 2,065,224 flows, respectively, from CICIDS2017 and UNSW-NB15 (see \autoref{fig:datasets}). For pre-processing, we eliminate duplicate instances and \emph{z}-normalize the data. We use a split of $\frac{2}{3}$ for training and $\frac{1}{3}$ for evaluation and testing.

\subsection{Performance}

In what follows, we show the performance of the EagerNet architecture taking into account various aspects. We stop training all models after 800 epochs to ensure fair comparison and we monitor both training and validation losses to prevent over-fitting.

\subsubsection{Comparability to \glspl{fcnn}}
\label{subsubsec:comparability_to_fnns}

In \autoref{tab:comparability}, we show the F1 score obtained at the last layer of two architectures: Simple \gls{fcnn} and EagerNet with the same number of layers and neurons. We use the F1 score to reduce the class imbalance bias. Results suggest that on average, EagerNet achieves similar scores to a conventional \gls{fcnn} at the last layer across all variants and datasets. It is worth mentioning, however, that EagerNet is used in this case as a conventional neural network and predictions are taken only from the last layer hence, comparing the influence of our training procedure on the typical workflow of the neural network. Using predictions from various layers by setting a confidence threshold might lead to higher accuracy if middle layers are optimized to predict certain classes.

\begin{table}[H]

\centering
\begin{tabular}{ccrc}
\toprule
\textbf{Dataset} & \textbf{Variant*} & \textbf{\gls{fcnn}} & \textbf{EagerNet} \\
\midrule
\multirow{2}{*}{CICIDS2017} & Binary & 0.989 & 0.993 \\
 & Multiclass & 0.979 & 0.920 \\
\midrule
\multirow{2}{*}{UNSW-NB15} & Binary & 0.919 & 0.908 \\
 & Multiclass & 0.882 & 0.880 \\
\midrule

\end{tabular}
\vspace{1ex}

{\raggedright * All architectures consist of 10 layers $\times$ 64 neurons in addition to input and output layers. \par}
\caption{F1 scores at the last layer of each architecture.}
\label{tab:comparability}
\end{table}

\subsubsection{Weighted loss}
In \autoref{tab:weights} we gather results from different EagerNet networks trained using different numbers of hidden layers and weight distributions each with 128 neurons per layer. In addition, we compare the results using four metrics (Accuracy, Precision, Recall and Youden's J) and show the performance of the last layer (by setting the confidence level to the maximum). The CICIDS2017 indices 
show that the weight distribution and the architecture itself (number of layers) have little effect on performance. In fact, since the last layer is always trained with one loss regardless of the weight distribution, only the first few layers are affected by the combined-loss learning procedure. Nonetheless, since the accuracy did not improve when increasing the number of layers, that implies that most network patterns are already separable after the few first layers which makes the classification task for CICIDS2017 manageable even for shallow neural networks. Results of UNSW-NB15 show a slightly different behavior. 
The uniform weighting of losses gives best scores on average and, in addition, deeper architectures outperform the shallow ones. This implies that the network patterns in this specific dataset are complex than in the previous one therefore, requiring more non-linearities (layers) to find the optimal \emph{input:prediction} mapping function.

\begin{table}[H]
\centering

\begin{tabular}{cccrrrr}
\toprule
\textbf{Variant} & \textbf{Weights} & \textbf{Layers} & \textbf{Acc.} & \textbf{Prec.} & \textbf{Rec.} & \textbf{J.}\\
\midrule
\multirow{6}{*}{\rotatebox{90}{CICIDS2017}} & \multirow{2}{*}{Uniform} & 5 & 0.996 & 0.996 & 0.990 & 0.989 \\
 & & 3 & 0.997 & 0.997 & 0.990 & 0.990 \\
 & \multirow{2}{*}{Increasing} & 5 & 0.997 & 0.998 & 0.990 & 0.989 \\
 & & 3 & 0.997 & 0.997 & 0.990 & 0.989 \\
 & \multirow{2}{*}{Decreasing} & 5 & 0.997 & 0.998 & 0.989 & 0.989 \\
 & & 3 & 0.997 & 0.997 & 0.990 & 0.990 \\
\midrule
\multirow{6}{*}{\rotatebox{90}{UNSW-NB15}} & \multirow{2}{*}{Uniform} & 5 & 0.988 & 0.828 & 0.867 & 0.860 \\
 & & 3 & 0.988 & 0.856 & 0.800 & 0.795 \\
 & \multirow{2}{*}{Increasing} & 5 & 0.988 & 0.853 & 0.821 & 0.816 \\
 & & 3 & 0.988 & 0.844 & 0.817 & 0.811 \\
 & \multirow{2}{*}{Decreasing} & 5 & 0.988 & 0.843 & 0.824 & 0.819 \\
 & & 3 & 0.988 & 0.851 & 0.819 & 0.813 \\

\end{tabular}

\vspace{1ex}

{\raggedright * All architectures consist of 128 neurons per layer.\par}
\caption{Effect of weights on the performance of different networks}
\label{tab:weights}

\end{table}

\subsubsection{Accuracy-Speed tradeoff}
\label{subsubsec:confidence-speed-tradeoff}

\begin{figure*}
\center
\begin{tabular}{cccc}
\subfloat[CICIDS2017 (3 layers $\times$ 128 neurons).]{\includegraphics[width=0.78\columnwidth]{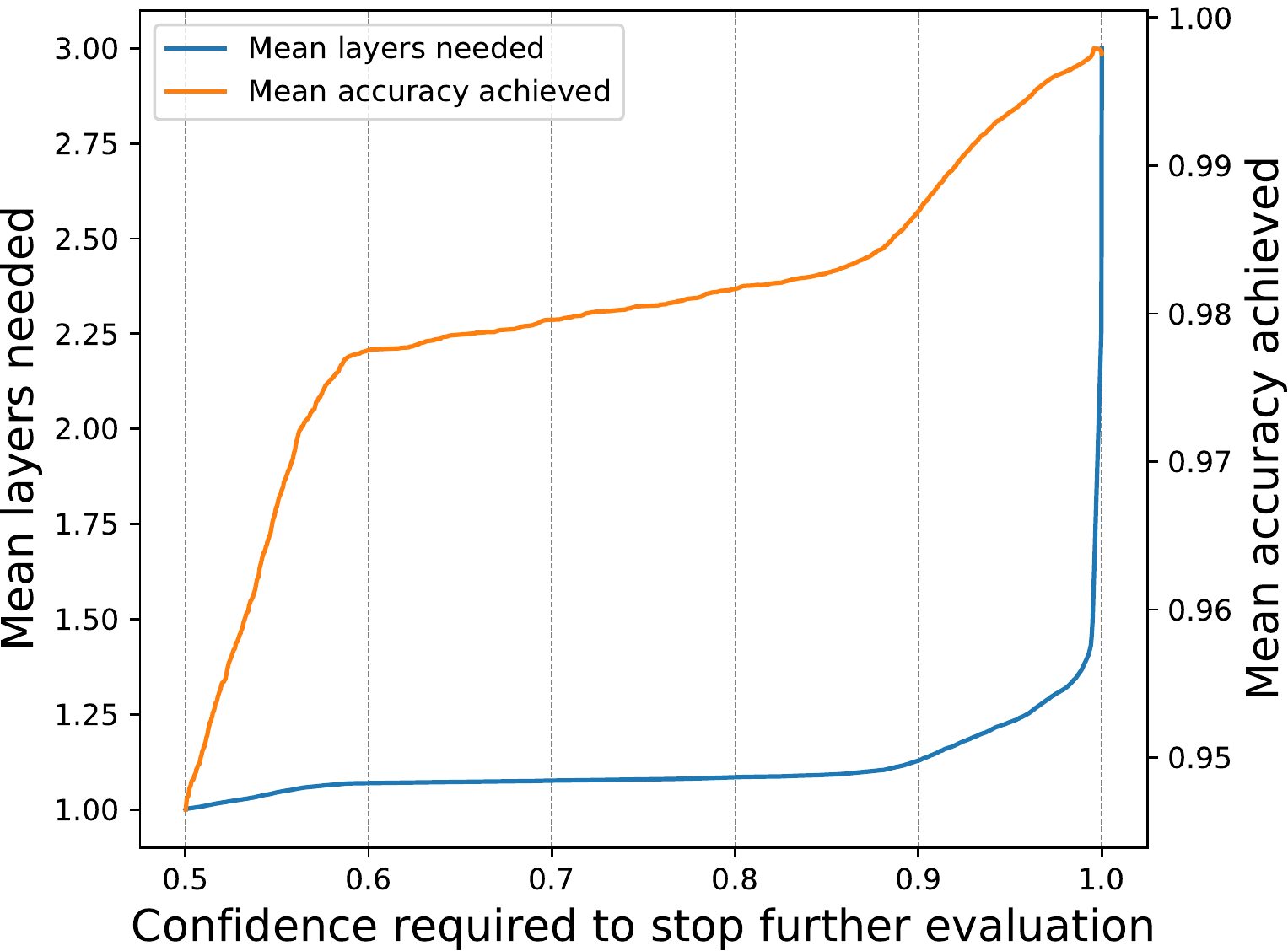}} &
\subfloat[CICIDS2017 (12 layers $\times$ 64 neurons).]{\includegraphics[width=0.78\columnwidth]{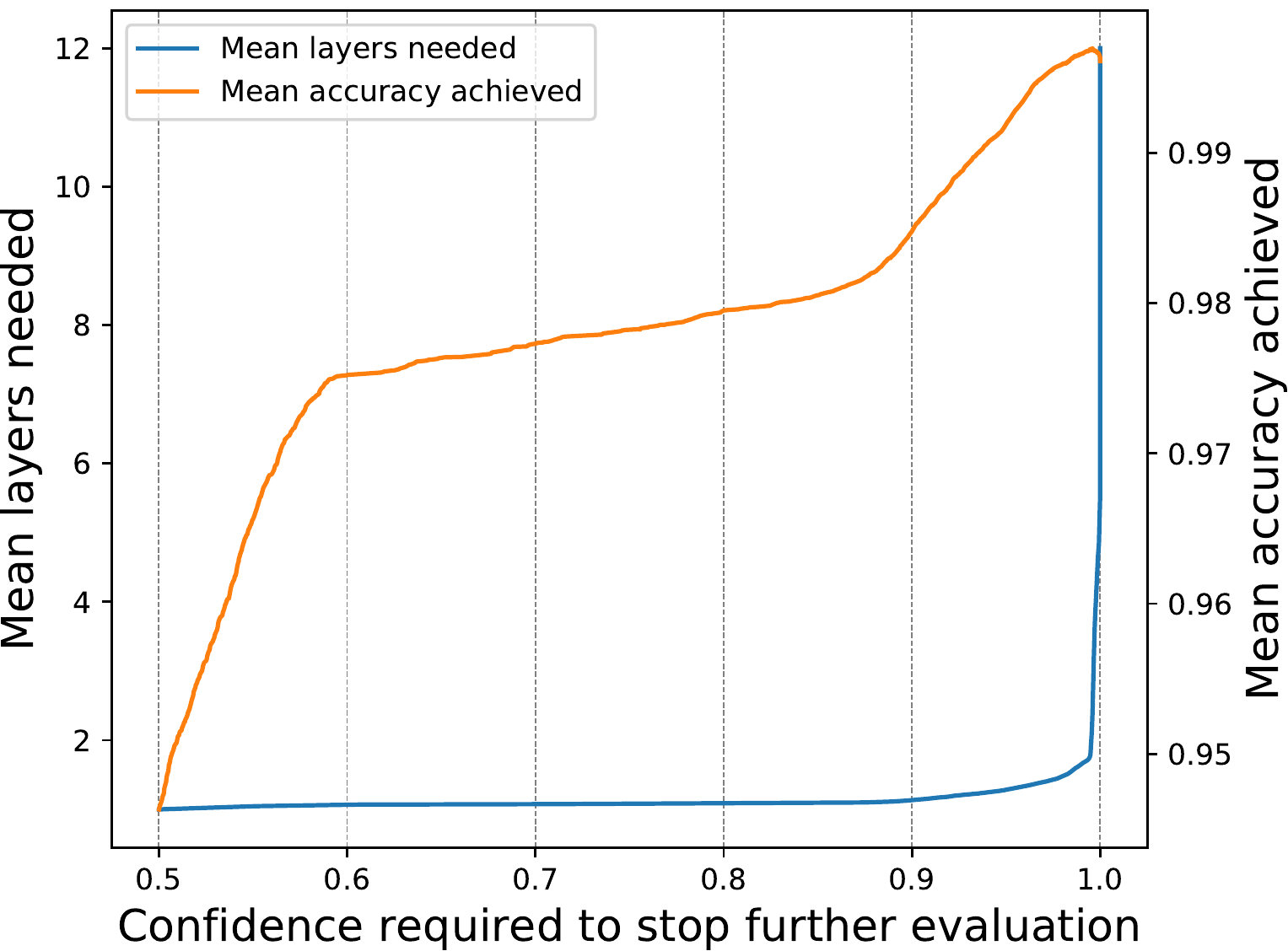}}\\
\subfloat[UNSW-NB15 (3 layers $\times$ 128 neurons).]{\includegraphics[width=0.78\columnwidth]{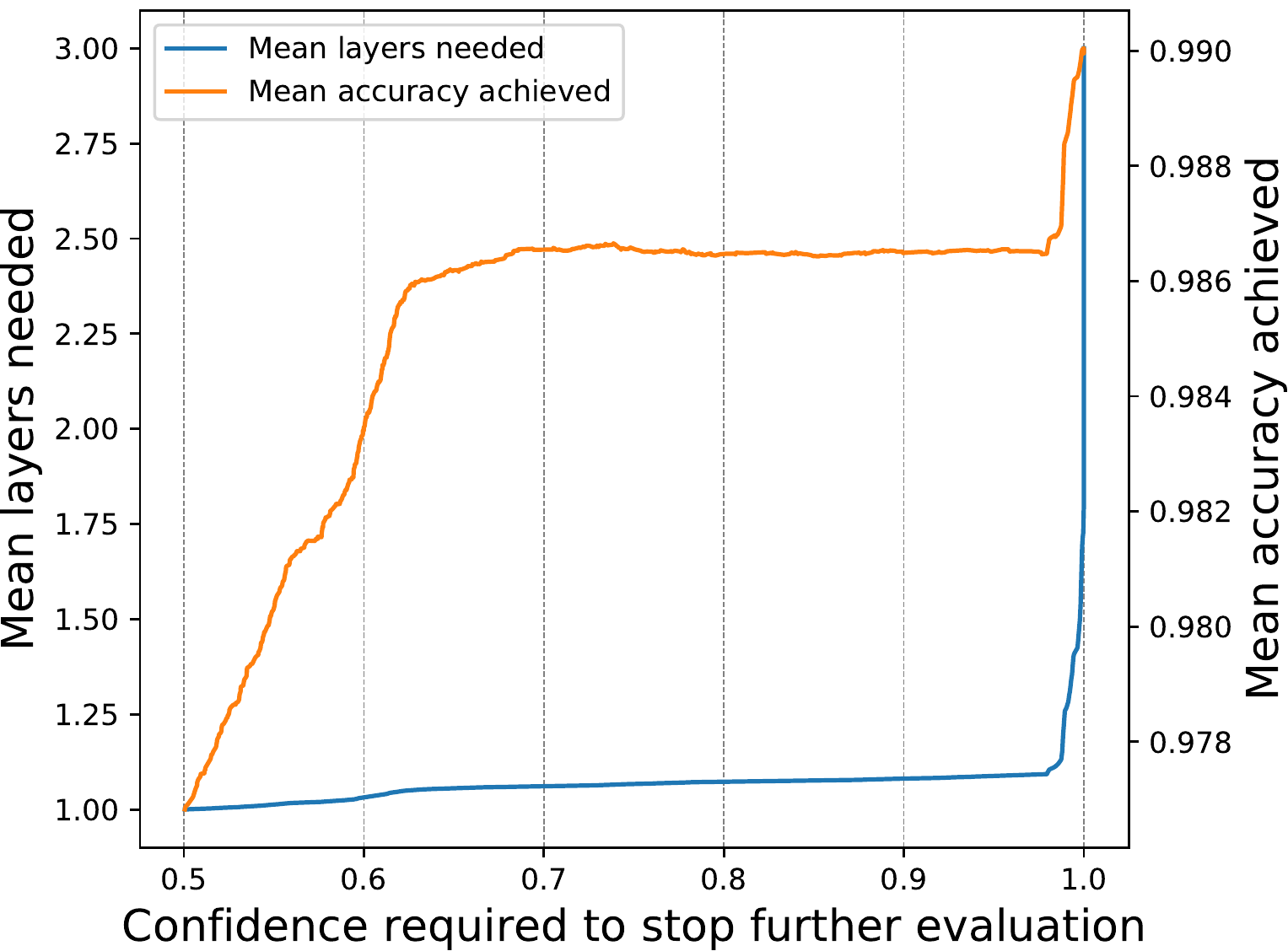}} &
\subfloat[UNSW-NB15 (12 layers $\times$ 64 neurons).]{\includegraphics[width=0.78\columnwidth]{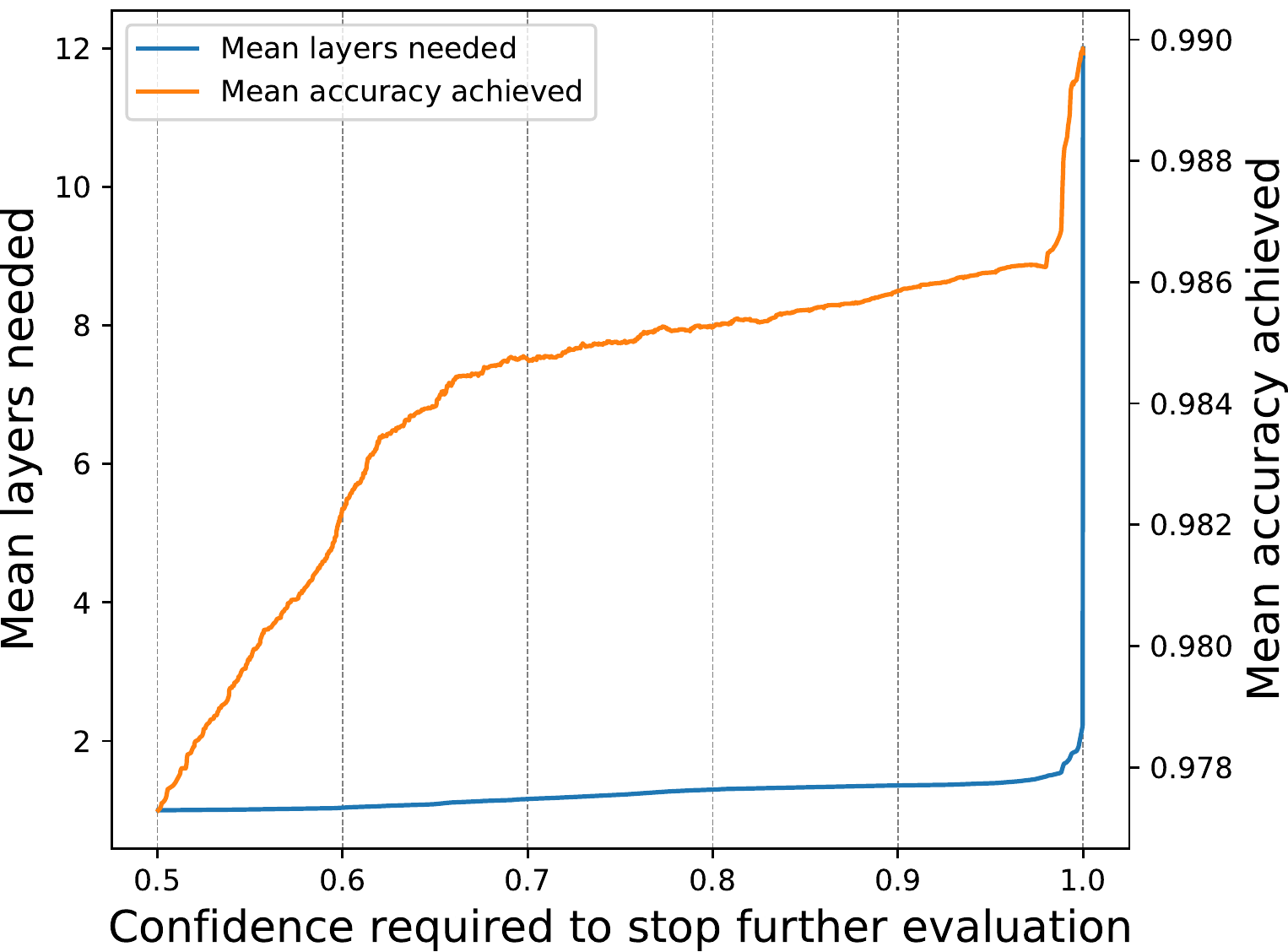}}
\end{tabular}
\caption{Confidence threshold effect on accuracy and number of needed layers. The maximum achieved accuracy does not represent the score obtained at the last layer but with a combination of best layers depending on the selection. Therefore, the last layer's accuracy alone might be less.}
\label{fig:confidence}
\end{figure*}

\autoref{fig:confidence} shows the accuracy-speed balance in terms of confidence threshold. Four different networks for both datasets are shown. We experiments with two networks: 128 neurons by 3 layers and 64 neurons by 12 layers. The confidence threshold is shown on the horizontal axis, the average number of layers used over all samples and the average accuracy achieved are both shown on the vertical axis. Curves show promising and consistent results, whereby the network always uses few layers for the majority of samples and only continues to evaluate further layers if the confidence level required is too high or the sample is noisy thus it is difficult to make correct decisions in early layers. We observe moreover two phenomena:
\begin{itemize}
\item The accuracy of CICIDS2017 increases almost linearly in sections with respect to confidence and reaches its maximum when all layers are evaluated.
\item The accuracy of UNSW-NB15 is stable when the confidence level is increased from $\approx 0.62$ to until $\approx 0.98$ where it leaps noticeably once a further layer is evaluated. This suggests that some patterns were ``learned'' only on that specific layer and hence, respective samples suddenly were correctly classified.
\end{itemize}
Overall, EagerNet confirms that it is possible to trade a tiny percentage of accuracy in order to save a significant amount of resources. In section \ref{subsubsec:comparability_to_fnns}, we see that, it is feasible to achieve comparable accuracies to traditional \gls{fcnn} allowing the reduction of costs.


\paragraph{Optimal Confidence Threshold}
During deployment, the optimum confidence threshold is set by the user and plays a role in the overall achieved accuracy. Setting the threshold too low can cause the network to make decisions at an early stage, thereby using fewer resources but achieving low accuracy. Similarly, setting the threshold too high can cause the network to always use all layers, which means a waste of resources. The desired confidence level can therefore be obtained after the training phase by determining on the basis of the desired accuracy as shown in \autoref{fig:confidence}.

\subsubsection{What did the network learn?}

\begin{figure*}
\center
\begin{tabular}{cccc}
\subfloat[CICIDS2017 (12 layers $\times$ 64 neurons).]{\includegraphics[width=0.52\columnwidth]{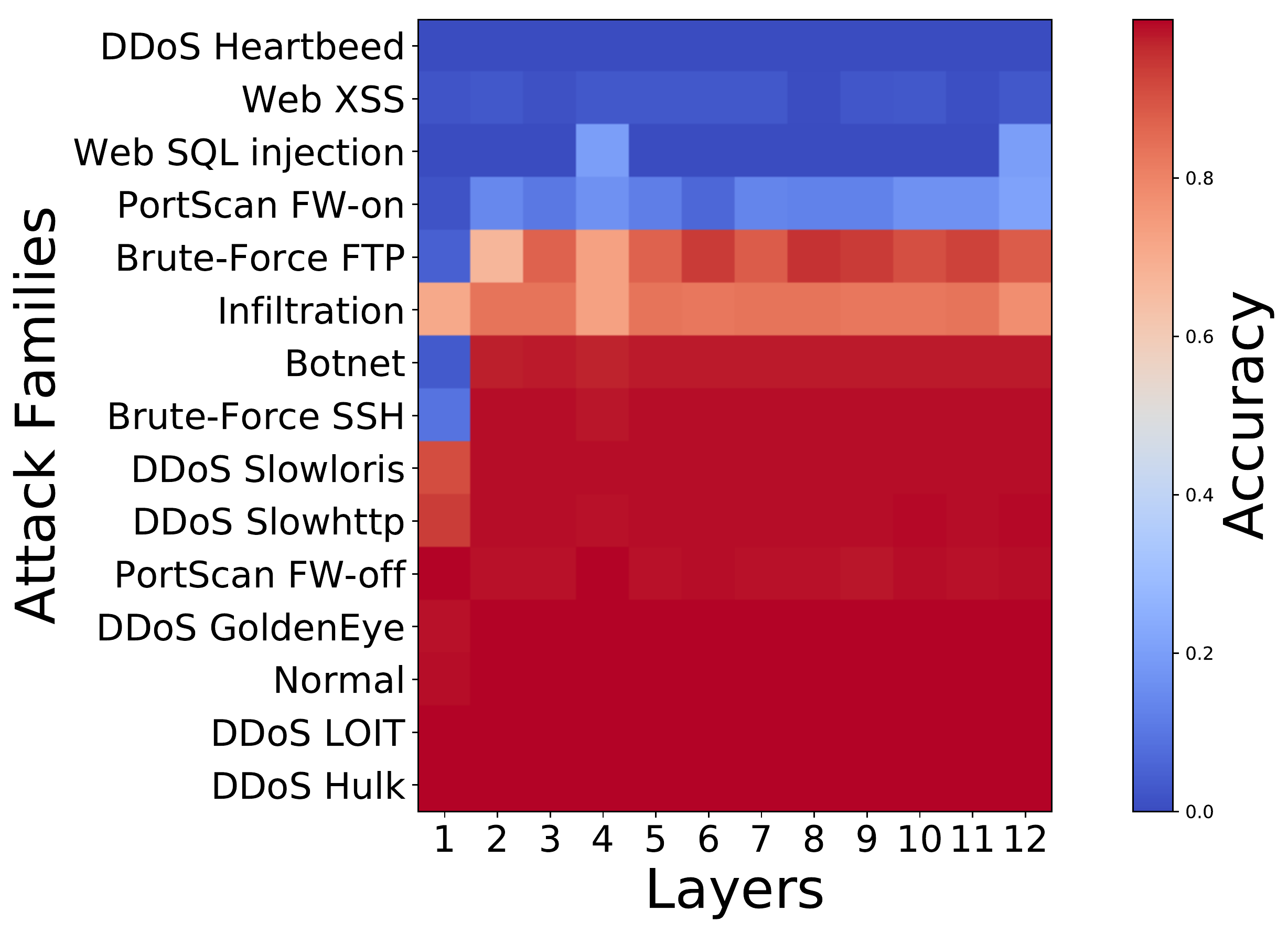}} &
\subfloat[CICIDS2017 (5 layers $\times$ 128 neurons).]{\includegraphics[width=0.37\columnwidth]{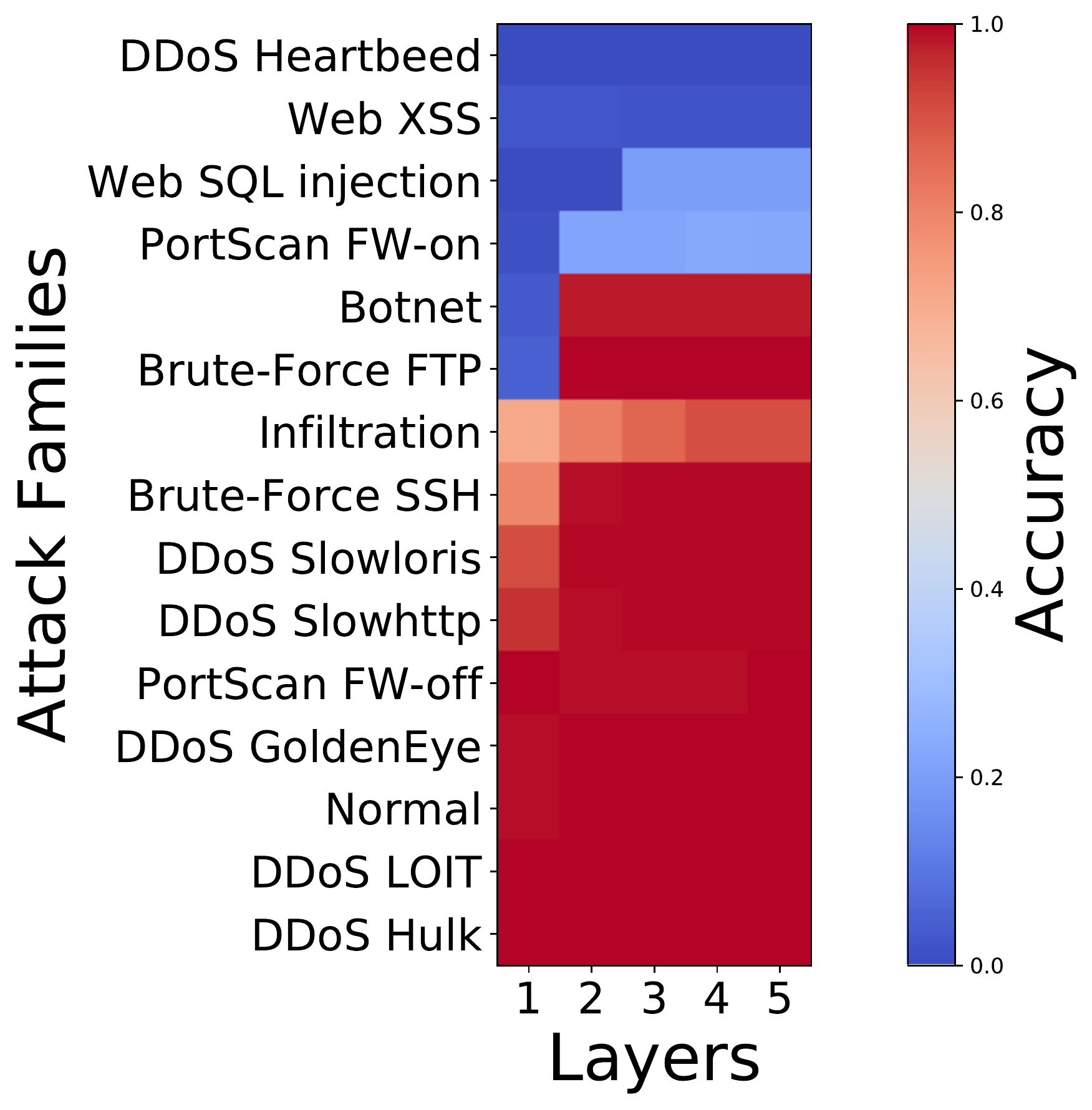}} 
\subfloat[UNSW-NB15 (12 layers $\times$ 64 neurons).]{\includegraphics[width=0.64\columnwidth]{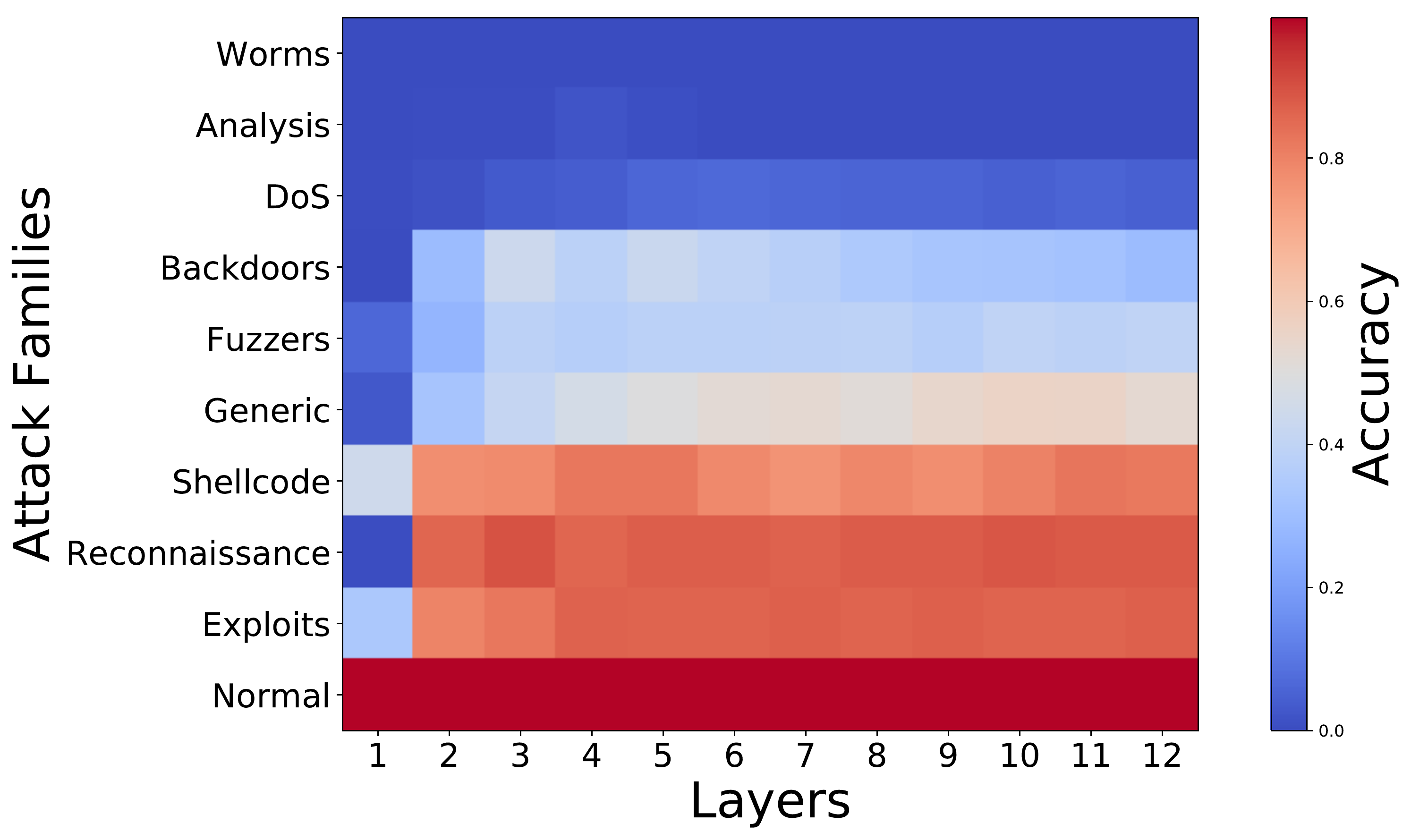}} &
\subfloat[UNSW-NB15 (5 layers $\times$ 128 neurons).]{\includegraphics[width=0.4\columnwidth]{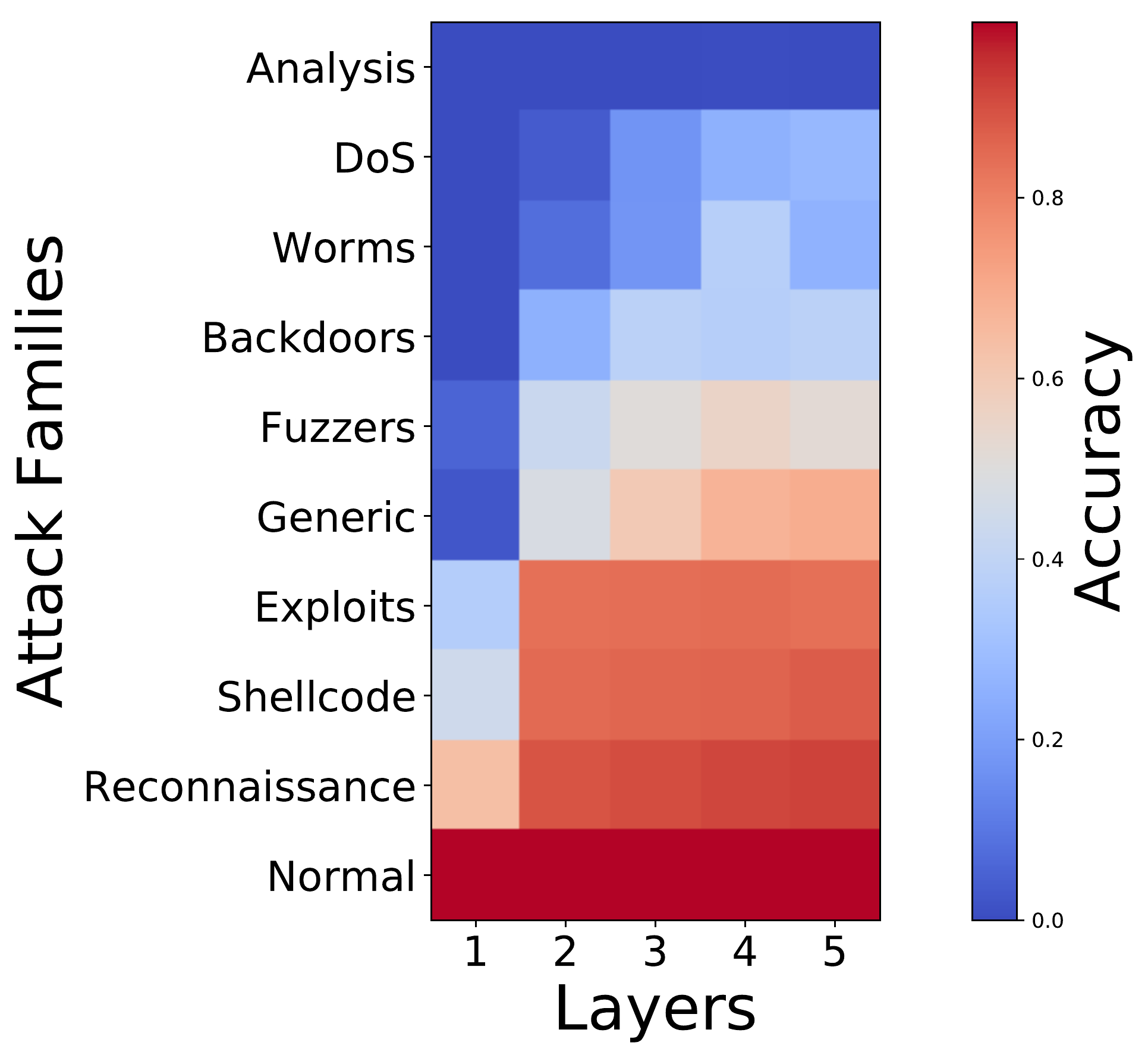}}
\end{tabular}
\caption{Accuracy achieved on a test set per layer and attack family (sorted vertically). Some attacks have poor accuracy because only few samples are observed (refer to \autoref{fig:datasets}). Forcing the network to learn those samples might lead to over-fitting. Additionally, the effect of over-sizing the neural network immediately appears and allows to apply the EagerNet strategy.}
\label{fig:multiclass}
\end{figure*}

In order to understand how EagerNet determines at which layer the best to stop evaluation, we use the multiclass architecture and conduct an explanatory analysis. In \autoref{fig:multiclass}, we show the accuracy of the prediction obtained by different networks per layer and attack family. We obtain two conclusions:
\begin{itemize}
\item  On average, the accuracy of predictions increases the deeper we get into the network. This was somehow expected, as more layers allow for more abstraction of the input space and thus better separation of instances. However, only few layers are needed to obtain the maximum accuracy per attack family. This implies that the EagerNet architecture significantly decreases resources when used for intrusion detection applications keeping the accuracy at its best.
\item  On the other hand, some attack families (Analysis, Worms, Web XSS, DDoS Heart-bleed etc.) show extremely poor accuracy. This is due to two reasons: (1) too few samples of a category are mostly ignored by the network to reduce over-fitting and (2) The loss optimized to learn such patterns at a specific layer is over-written by the loss that is optimized to learn the majority classes hence erasing this ``fragile'' knowledge.
\end{itemize}

\subsubsection{Advantage of backpropagating the losses of all outputs until the beginning}

Instead of backpropagating each loss through all layers, it is also possible to only backpropagate each loss of the intermediate layers only to the previous layer but not until the beginning. Only the last layer's output's loss would then be backpropagated until the beginning but the other intermediate output's losses only one step. One benefit of this would be to save computational resources during training. Our results showed that this method results in worse accuracy for the intermediate layers' outputs. Backpropagating all losses at once is thus important and helps to update the weights of the network at the same time as adjusting the weights of the output neurons, thereby allowing the neural network to distribute the knowledge across all weights and make accurate predictions at each layer.

\section{Conclusion}

This paper explores if it is viable to terminate the evaluation of \glspl{fcnn} before the final layer to save computational resources. We propose EagerNet, an architecture that allows predictions to be made as soon as the neural network is sufficiently confident, saving energy and resources and making it possible to implement similar architectures in real-time applications where prediction speed is relevant. We evaluate our approach using two intrusion detection datasets and obtain satisfactory results indicating the possibility of achieving comparable evaluation scores to traditional \glspl{fcnn} yet using only a segment of the neural network, as most network traffic flows are simpler to classify and only a minority need deeper propagation through the neural network.
Furthermore, we show that by setting a confidence threshold during deployment, it becomes possible to trade accuracy for resources usage and energy consumption. The EagerNet architecture is thus characterized by its simplicity and effectiveness, making it an ideal solution for potential \gls{fcnn}-based \glspl{ids} whenever resources are of primary concern.

\section*{Acknowledgements}
The Titan Xp used for this research was donated by the NVIDIA Corporation.

\bibliographystyle{IEEEtran}
\bibliography{biblio}

\end{document}